\documentclass{scrartcl}

\usepackage[utf8]{inputenc}
\usepackage{amsmath, amssymb, amsthm}
\usepackage{pgffor}
\usepackage{float}
\usepackage{geometry}
\usepackage{float}
\usepackage{xcolor}
\usepackage{hyperref}
\usepackage{graphicx}
\usepackage{makecell}
\usepackage{array}
\usepackage{xspace}
\usepackage{cleveref}
\usepackage{caption}
\usepackage{placeins}
\usepackage[normalem]{ulem} 
\usepackage{booktabs} 

\usepackage[linecolor=white,backgroundcolor=white,bordercolor=white,textsize=tiny]{todonotes}
\let\todon\todo
\renewcommand{\todo}[1]{\todon{\color{magenta}#1}}

\usepackage{mathrsfs}
\newcommand\TreeAlgo{\ensuremath{\mathsf{Tree_{\,5/3}}}}

\newcommand{\rawGPE}{\widehat{\mathsf{GPE}}}
\newcommand{\GPE}{\ifmmode\mathsf{GPE}\else\textsf{GPE}\fi\xspace}
\newcommand{\rawCTPE}{\widehat{\mathsf{CTPE}}}
\newcommand{\rawPE}{\widehat{\mathsf{PE}}}
\newcommand{\PE}{\ifmmode\mathsf{PE}\else\textsf{PE}\fi\xspace}
\newcommand{\perm}[1]{%
  \textcolor{black}{%
  \mathtt{%
  [%
    \StrLen{#1}[\stringLength]
    \foreach \n in {1,...,\stringLength}{%
      \StrChar{#1}{\n}
      \ifnum\n<\stringLength
      \,
      \fi
    }
  ]
  }
}
}

\newcommand\windowsize{\mathsf{window\_size}\xspace}

\usepackage[style=alphabetic]{biblatex}
\usepackage{wrapfig}  
\usepackage{authblk}
\title{Global Permutation Entropy}
\author[1]{Abhijeet Avhale}
\author[1]{Joscha Diehl}
\author[1]{Niraj Velankar}
\author[1]{Emanuele Verri}
\affil[1]{\footnotesize Institute of Mathematics and Computer Science, University of Greifswald, 17489 Greifswald, Germany}
\date{\today}

\newcommand\DEF[1]{\textbf{#1}}

\begin{document}
\maketitle
\begin{abstract}
    Permutation Entropy, introduced by Bandt and Pompe, is a widely used
    complexity measure for real-valued time series
    that is based on the relative order of values within
    \emph{consecutive} segments of fixed length.
    After standardizing each segment to a
    permutation and computing the frequency distribution of these permutations,
    Shannon Entropy is then applied to quantify the series' complexity.

    We introduce Global Permutation Entropy (\GPE), a novel
    index that considers \emph{all} possible patterns of a given length,
    including non-consecutive ones.
    Its computation relies on recently developed algorithms that enable the
    efficient extraction of full permutation profiles. We illustrate some
    properties of \GPE and demonstrate its effectiveness through experiments on
    synthetic datasets, showing that it reveals structural
    information not accessible through standard permutation entropy.
    We provide a Julia package for the calculation of \GPE at
    \url{https://github.com/AThreeH1/Global-Permutation-Entropy}.
\end{abstract}

\tableofcontents

\section{Introduction}

\textbf{Permutation Entropy} (\PE), introduced by Bandt and Pompe in \cite{bandt2002permutation}, is a simple and widely used method for quantifying the complexity of time series. Given a time series, classical \PE\ analyzes ordinal patterns extracted from short consecutive segments of the time series by ranking their values and counting the frequencies of the resulting permutations. The Shannon entropy of this empirical distribution then quantifies the diversity of these ordinal patterns: higher entropy reflects greater randomness or complexity in the data.

The computational cost of \PE\ scales linearly with the length of the time series, \(\mathcal{O}(n)\), making it
applicable even to large datasets. Moreover, since \PE\ depends solely on the
relative ordering of values, it is invariant under monotonic
transformations and robust to observational noise. 
Often, a delay parameter \(\tau\) is introduced (\cite{keller2003symbolic}) to sample points within the time series at specific time steps, enabling the capture of slower dynamics or multi-scale structures.
$\PE$ has been successfully applied to data in various fields, including finance \cite{zhao2013measuring},
neuroscience \cite{ferlazzo2014permutation,li2007predictability},
and machine learning \cite{liu2022scinet}.

Since classical permutation entropy restricts attention to patterns formed from consecutive or regularly spaced points, it potentially overlooks more intricate dependencies, especially in short or structured signals. 
We therefore propose \textbf{Global Permutation Entropy}
(\GPE), whose computation is based on \emph{all} strictly increasing index
combinations of size \(k\) drawn from the full time series \((X_t)_{t=1}^n\). 
In other words, \GPE requires the full \emph{permutation profile} of order
\(k\), which records the frequency of each permutation of size \(k\) across all
\(\binom{n}{k}\) increasing subsets of the index set.
Naively, this has a computational cost of \(\mathcal{O}(n^k)\), which becomes
prohibitive for large \(n\).  However, recent algorithmic breakthroughs have led
to efficient methods for counting permutation patterns, with complexities that
scale at most quadratically\footnote{modulo a polylogarithmic factor.} in \(n\)
for all orders up to \(k = 7\)~\cite{beniamini2024counting, diehl2024efficient,
even2021counting}. See Section~\ref{subsec:fast_prof} for details. As a result,
computing the \GPE\ for orders up to 7 is now computationally feasible, even for
large input sizes.


\subsection{Our Contributions}

\begin{itemize}
    \item We propose \emph{Global Permutation Entropy} (\GPE), a new complexity measure that is conceptually simple but whose naive calculation is computationally intractable for even moderate window sizes and orders. Only recent advances in permutation pattern counting have made its practical implementation viable.

    \item We provide a Julia library (\url{https://github.com/AThreeH1/Global-Permutation-Entropy}) for computing global permutation entropy
    up to order~6, using corner trees~\cite{even2021counting} and their generalizations~\cite{beniamini2024counting,diehl2024efficient} to count global permutation patterns. In particular, we include fast implementations for the 3-, 4-, 5-, and 6-permutation pattern profiles.

    \item We test the library on synthetic data to tease out differences between \GPE and \PE.

    \item Despite its higher computational cost compared to classical \PE, \GPE\ offers several desirable properties:
    \begin{itemize}
        \item For completely random signals and fixed order $k > 2$, $\GPE(k)$ converges faster than $\PE(k)$ to the ``correct'' value of~$1$. See Section~\ref{subsection:windows_size_convergence}.
        \item In certain scenarios involving a noisy periodic signal with a sudden increase in noise level, \GPE\ detects changes more quickly than \PE. See Section~\ref{subsection:noise_detection}.
        \item When noise is gradually added to a periodic signal, \GPE\ remains highly descriptive and robust across different orders and window sizes. Its performance is also comparable to \PE\ when the latter is carefully tuned (e.g., by using optimal delays or averaging across delays). See Section~\ref{subsection:periodic_additive_noise_linear_increase}.
    \end{itemize}
\end{itemize}

\section{Global Permutation Entropy}

Let \((X_t)_{t=1}^n\) be a real-valued time series of length \(n\). For each strictly increasing index tuple \((i_1, i_2, \ldots, i_k)\) with \(1 \le i_1 < i_2 < \cdots < i_k \le n\), we extract the subsequence \((X_{i_1}, X_{i_2}, \ldots, X_{i_k})\) and rank its values from smallest to largest. This ranking\footnote{In case a tie occurs, values are ordered according to their time of appearance.} defines a unique permutation \(\sigma \in S_k\), where \(\sigma(j) = r\) means that the element in position \(j\) of the subsequence is the \(r\)-th smallest among the \(k\) elements.  
We denote the collection of all strictly increasing index tuples of size \(k\) by
\[
\mathcal{I}_k := \{(i_1, i_2, \ldots, i_k) \mid 1 \le i_1 < i_2 < \cdots < i_k \le n \}.
\]

The relative frequency of each ordinal pattern \(\sigma \in S_k\) is then given by
\begin{align}
    \label{eq:p_sigma}
    p(\sigma) := \frac{\#\{(i_1, \ldots, i_k) \in \mathcal{I}_k : (X_{i_1}, \ldots, X_{i_k}) \text{ has permutation type } \sigma \}}{\binom{n}{k}}.
\end{align}

The (raw) \textbf{Global Permutation Entropy} (GPE) of order \(k\) is defined as the Shannon entropy of this empirical distribution:
\begin{align*}
    \rawGPE(k) := -\sum_{\sigma \in S_k} p(\sigma) \log p(\sigma),
\end{align*}
with the convention \(0 \log 0 := 0\).

For completeness, recall that the (raw) \textbf{Permutation Entropy} (PE) of order \(k\) with delay $\tau$ is defined as
\begin{align*}
  \rawPE(k;\tau) &:= -\sum_{\sigma \in S_k} q(\sigma) \log q(\sigma), \\
  q(\sigma) &:= \frac{\#\{ 1 \le i \le n - \tau(k -1) : (X_{i},X_{i+\tau}, \ldots, X_{i+\tau(k-1)}) \text{ has permutation type } \sigma \}}{ n - \tau(k -1)}.
\end{align*}
Observe that
\begin{itemize}
    \item For any order \(k\), both $\rawGPE(k)$ and $\rawPE(k;\tau)$ take values in the interval $[0,\, \log(k!)]$.
        The lower bound is attained when only a single pattern occurs, namely either the identity (increasing) or the reverse (descending) permutation, as in monotone signals, while the upper bound is reached when all $k!$ patterns are equally likely (as in purely random signals).

    \item Both $\GPE$ and $\PE$ depend only on the relative ordering of the values. They are therefore \emph{invariant under monotone transformations} of the time series and \emph{robust to observational noise}.  

\end{itemize}

In what follows, we shall work with the normalized versions of both quantities:
\[
\GPE(k) := \frac{\rawGPE(k)}{\log(k!)}, 
\qquad 
\PE(k,\tau) := \frac{\rawPE(k;\tau)}{\log(k!)}.
\]
When the default delay $\tau=1$ is used, we simply write $\rawPE(k) := \rawPE(k;1)$ and 
\(\PE(k) := \rawPE(k)/\log(k!).\)

\begin{figure}[H]
    \centering
    \includegraphics[scale=0.18]{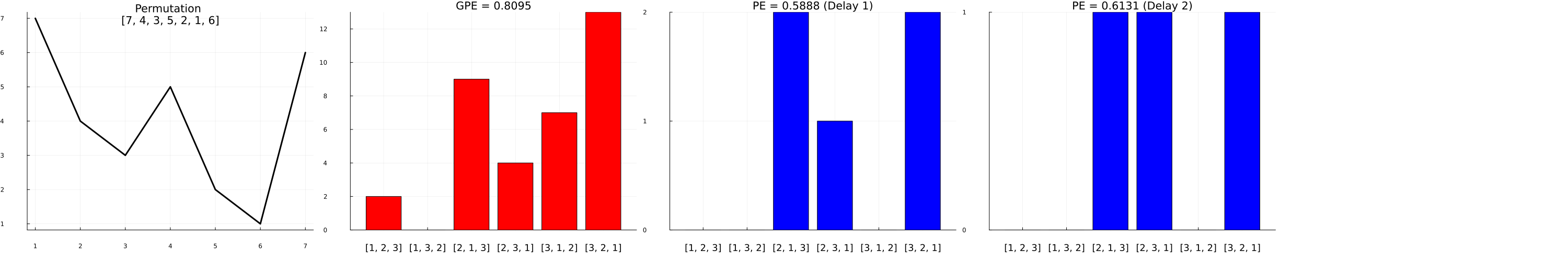}
    \caption{Example illustrating the 3-profile and pattern frequencies used to compute \(\GPE(3)\), \(\PE(3;1)\) and \(\PE(3;2)\) respectively.}
    \label{figure:example}
    \end{figure}

As an example, Figure~\ref{figure:example} illustrates the 3-profile of the permutation $[\mathtt{7\,4 \, 3 \, 5\, 2\, 1\, 6}]$ used to compute
$\GPE(3)$, along with the frequencies of the 3 patterns used to compute
$\PE(3;1)$ and $\PE(3;2)$ respectively. For the rest of the paper, we will stick to the convention that \(\GPE\) (and related quantities) will always be depicted in red, and \(\PE\) (and related quantities) in blue.

\begin{figure}[H]
    \centering
    \begin{minipage}[b]{0.45\linewidth}
        \centering
        \includegraphics[width=\linewidth]{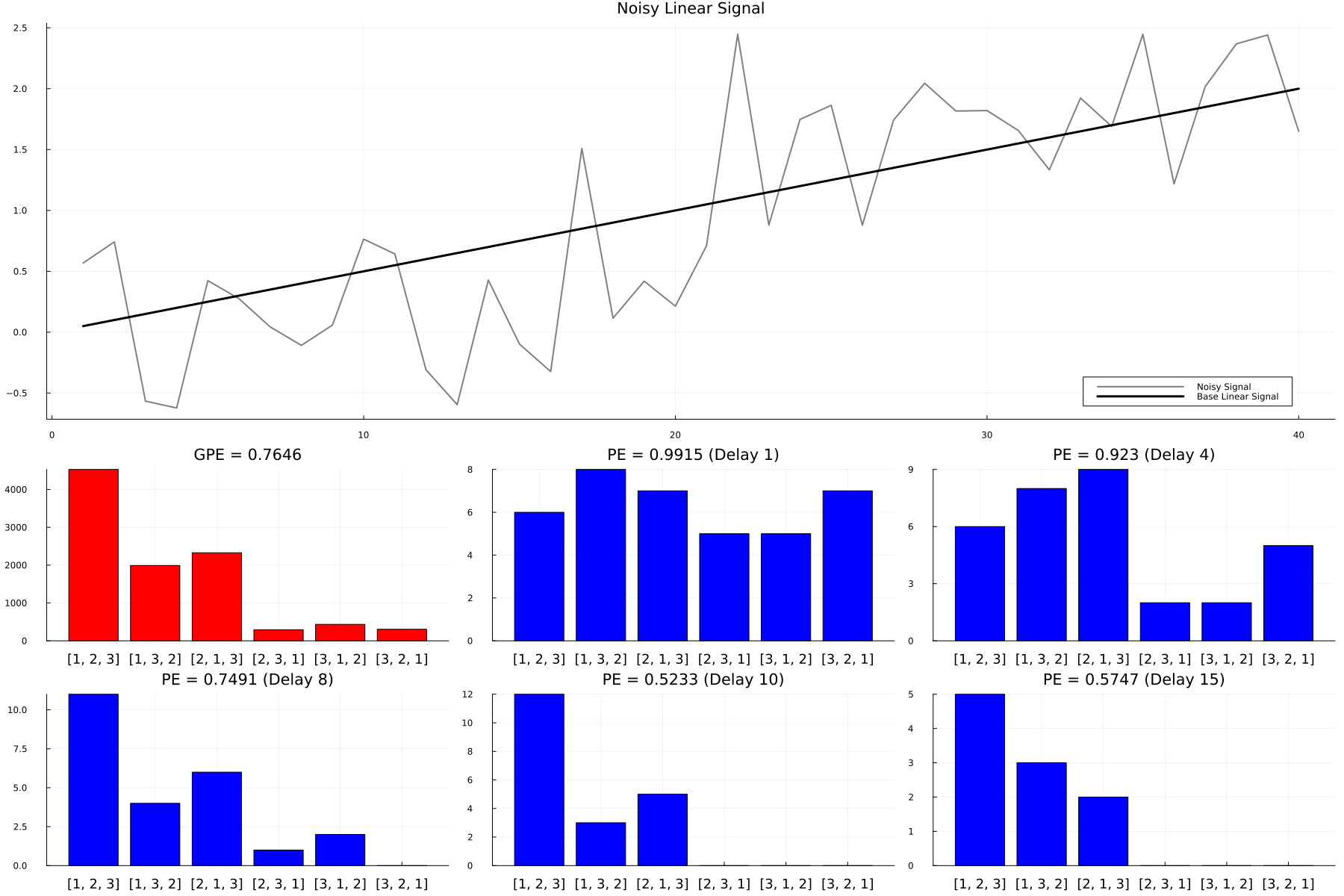}
    \end{minipage}
    \hspace{0.05\linewidth}
    \begin{minipage}[b]{0.45\linewidth}
        \centering
    \includegraphics[width=\linewidth]{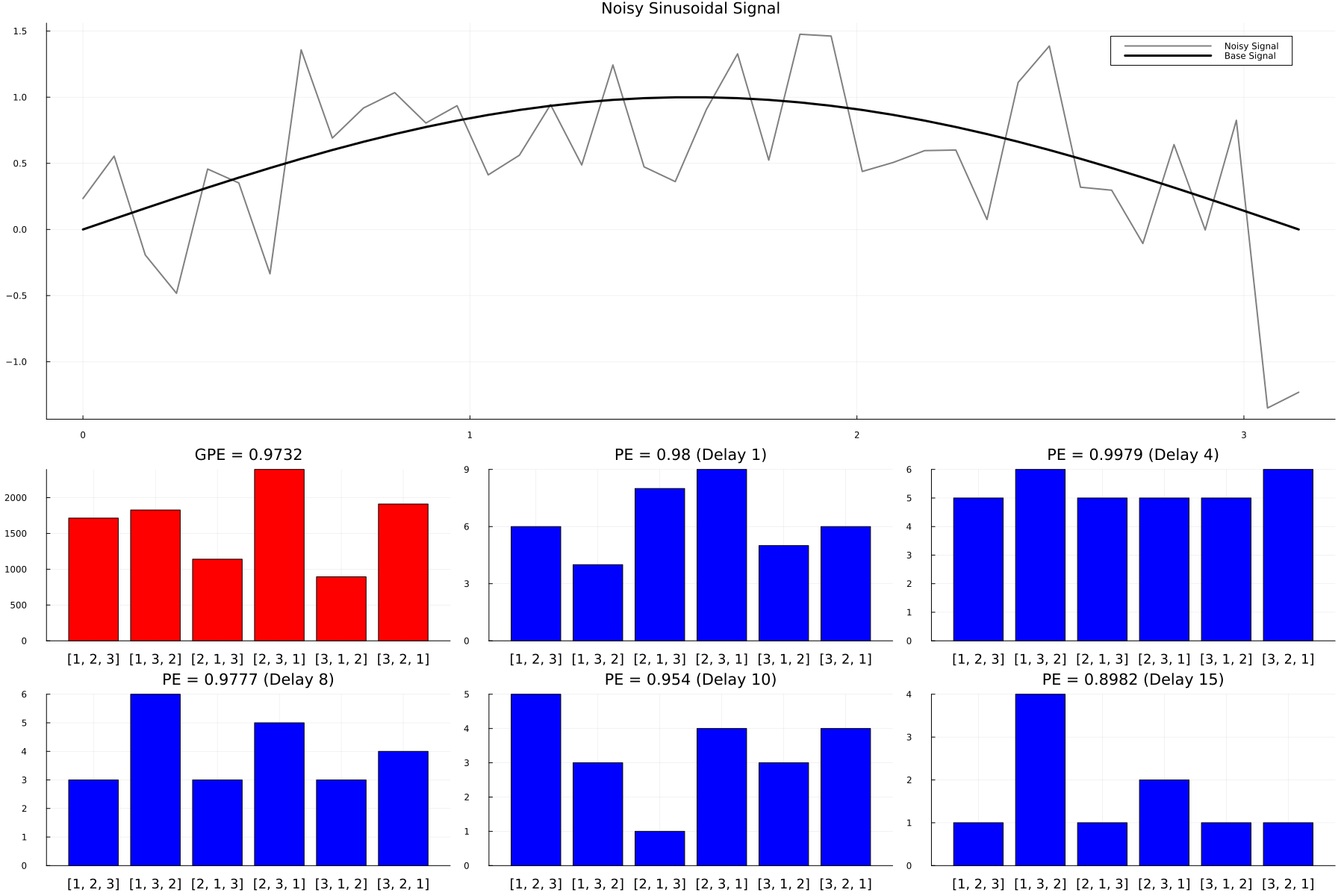}
   \end{minipage}
   \caption{Left: increasing straight line with noise. Right: Sinus function from 0 to $\pi$ with noise.}
  \label{figure:sine-comparison}
\end{figure}

In Figure~\ref{figure:sine-comparison}, on the left, we show the realization of a signal which consists of a straight line (with slope equal to 0.05) from which points 1 up to 40 are considered. At each time point, the added noise follows a normal distribution $\mathcal{N}(0,0.025)$. On the right, we depict the sinus function on 40 equispaced points between 0 and $\pi$. Again, at each time point, we add a realization from $\mathcal{N}(0,0.025)$.


\subsection{Fast Computation of the 2-,3-,4-,5-,6-Profiles}
\label{subsec:fast_prof}

In the seminal work \cite{even2021counting}, 
it was realized that (certain) permutation patterns can be counted
by counting combinatorial structures called \emph{corner trees}.
The tree structure of the latter leads to a fast algorithm for counting
their occurrences in a time series.

In order $2$ and $3$, \emph{all} permutation patterns can be counted
using corner trees, and the corresponding algorithm runs in time
\(\mathcal{O}(n \log n)\).
For higher orders,
additional patterns need to be counted.
Some have already been proposed in \cite{even2021counting},
with slight extensions in \cite{diehl2024efficient}.
Finally, with the work of \cite{beniamini2024counting}, all patterns up to order \(5\) can be counted in (almost) subquadratic time, while all patterns up to order \(7\) can be counted in (almost) quadratic time.


The \DEF{permutation profile} of order $k$ for a time series of length $n$ is the collection of counts of all permutation patterns of order $k$ appearing in the time series.
See \Cref{figure:example} for the $3$-profile for a time series of length $7$.
We remark that when using the methods described above to compute the permutation profile
of order $k$ one necessarily computes the profiles for all orders \(i \leq k\) as well.

\paragraph{Profile Implementations}
In our repository~\cite{avhale2025global},
the profiles of order \( k = 2, \ldots, 6 \) are implemented.
For details regarding the implementation and time-complexity of the algorithms, we refer to \cite{beniamini2024counting, even2021counting, diehl2024efficient}. We summarize them below:
\begin{itemize}
    \item \textbf{2-profile:}  
    Implemented using two \emph{corner trees}, each with two vertices. Corner trees were introduced in~\cite{even2021counting}. Counting the occurrences of these trees requires \( \mathcal{O}(n \log n) \) time, so the entire profile can be computed in \( \mathcal{O}(n \log n) \) time.

    \item \textbf{3-profile:}  
    Builds on the two-vertex corner trees from the 2-profile by adding six corner trees with three vertices.
    They span the full 3-profile and hence, the total computation time remains \( \mathcal{O}(n \log n) \).

    \item \textbf{4-profile:}  
    Extends the previous profiles by incorporating 23 additional corner trees with four vertices and the permutation pattern $[\mathtt{3\,2 \, 1 \, 4}]$, which is counted in \( \mathcal{O}(n^{5/3} \log^2 n) \) time using the algorithm from~\cite{even2021counting}.
    Thus, the entire 4-profile can be computed in \( \mathcal{O}(n^{5/3} \log^2 n) \) time.


\item \textbf{5-profile:} Extends the 4-profile with 100 corner trees, 10 tree double posets from the $\TreeAlgo$ family (introduced in~\cite{diehl2024efficient}) together with their images under nontrivial $D_4$ actions, all with five vertices. The remaining 10 double posets are obtained using \emph{marked patterns} (introduced in~\cite{beniamini2024counting}), also under the action of $D_4$.  The $\TreeAlgo$ elements generalize the counting of $[\mathtt{3\,2\,1\,4}]$ and can be computed in $\mathcal{O}(n^{5/3} \log^2 n)$ time. Among the 10 directions obtained using the marked patterns, two are counted in $\mathcal{O}(n^{5/3} \log^2 n)$ time, by extending the marked pattern $[\mathtt{3\,2\,1\,\underline{4}}]$ to length 5, yielding six double posets under the action of $D_4$, while one direction arises from counting occurrences of the pattern
$[\mathtt{4\,3\,2\,1\,{5}}]$ explicitely in $\mathcal{O}(n^{7/4} \log^2 n)$ time, yielding four double posets under the action of $D_4$. As the latter case dominates, the full 5-profile is computable in $\mathcal{O}(n^{7/4} \log^2 n)$ time.

\item \textbf{6-profile:} Extends the 5-profile with 463 corner trees, 44 tree double posets originating from the family $\TreeAlgo$, and 213 \emph{pattern trees} (generalisation of corner trees, introduced in~\cite{beniamini2024counting}), all with six vertices. 
The 44 tree double posets include elements of $\TreeAlgo$ as well as images under nontrivial $D_4$ actions. The 213 pattern trees can be counted in $\mathcal{O}(n^2 \log^4 n)$ time~\cite{beniamini2024counting}, which dominates the computation. 
Thus, the complete 6-profile is computable in $\mathcal{O}(n^2 \log^4 n)$ time.

\end{itemize}

\section{Experiments}

\subsection*{Methodology}

\textbf{Sliding window} Instead of computing \PE over the entire time series,
people often use a \emph{sliding window} approach, which was already proposed in \cite{bandt2002permutation}.
Sliding windows allow for a more localized analysis, often revealing transient
changes in the system's behavior that might be missed when considering the full
sequence at once.
For that reason, this approach is also useful for \GPE,
in particular, since \GPE considers \emph{all} possible ordinal patterns,
the entropy of the \emph{entire} sequence will often be close to 1,
independent of the underlying dynamics.

Let $k$ be a fixed order. At each time point $t$ (i.e. using a \emph{stride} of $1$; see \cite{little2017variance} for other choices), we select the window of data points
\[
[X_{t - \windowsize + 1}, \ldots, X_t]
\]
and compute either $\PE(k;\tau)$, or $\GPE(k)$.
Given a time series \((X_t)_{t=1}^n\), this yields a new time series of entropy
values \((Y_t)_{t=\windowsize}^n\), where each $Y_t$ represents the
entropy computed from the window ending at time $t$.
We do emphasize that inside a window, \PE
considers only consecutive patterns, while \GPE considers
all patterns.

\bigskip

\textbf{Choice of window size for} \GPE The window size should be chosen with an appropriate sample size in mind. For a reliable estimation of $\GPE$, we need to have a sample size
\begin{align}
  \label{eq:GPE_sample_size}
  \binom{\windowsize}{k}\gg k!
\end{align}
much larger than the number of patterns of order $k$.
The window size is varied over a reasonable range, 
from around the first value satisfying \eqref{eq:GPE_sample_size}
up to half the signal length.
In general, a window size that is too small will not capture enough
information about the signal, while a window size that is too large will
dilute the information by averaging over too many points. 

For data which is expected, or known, to have an underlying \emph{periodic structure}, the following approach has proven useful.
We compute the entropy values over time by sliding the window along the signal and then averaging these values, 
obtaining an averaged value of the windowed entropy as a function of the window size.
The curve is expected to have a minimum at around half the period of the signal,
since in this case the window (sometimes) covers (approximately) increasing or decreasing
windows of the signal.
If the curve has the shape of a typical curve in Figure~\ref{fig:entropy_average},
the minimum is expected to be near half the period of the signal.
In Section \ref{subsection:periodic_additive_noise_linear_increase}, we observe that selecting a window size between the minimum and twice this value (the expected period) yields good results. Furthermore, in Section \ref{subsection:noise_detection}, the best performance for \GPE is achieved when the window length is nearly equal to the period.

\bigskip
\textbf{Choice of window size for} \PE
For \PE, the effective sample size is reduced by the delay and order, and is given by
\[
\windowsize - \tau (k - 1).
\]
And again, it needs to be much larger than $k!$. 
The literature seems to be devoid of a principled approach for selecting the window size.
For our experiments, we try several window sizes.

\bigskip
\textbf{Delay for} \PE
In the literature (\cite[Section 4]{bandt2017new}) it is often recommended 
to average the delay parameter over a range of values, i.e.,
\begin{align*}
\overline{\PE}(k):=\frac{1}{|\mathcal{T}|} \sum_{\tau \in \mathcal{T}} \PE(k, \tau),
\end{align*}
where $\mathcal{T}$ is the chosen set of delay values.
Additionally, we also consider certain fixed delays.

\bigskip
\textbf{Order}
Both entropies depend on the order \(k\).
Regarding \PE, it is common to 
try orders in the range of $[3,7]$ (\cite{bandt2002permutation}).
Regarding \GPE, we usually try all currently feasible orders, i.e. \(k = 2, 3, 4, 5, 6\).

\subsection{Window size and convergence}
\label{subsection:windows_size_convergence}

\begin{figure}[H]

    \includegraphics[width=0.3\linewidth]{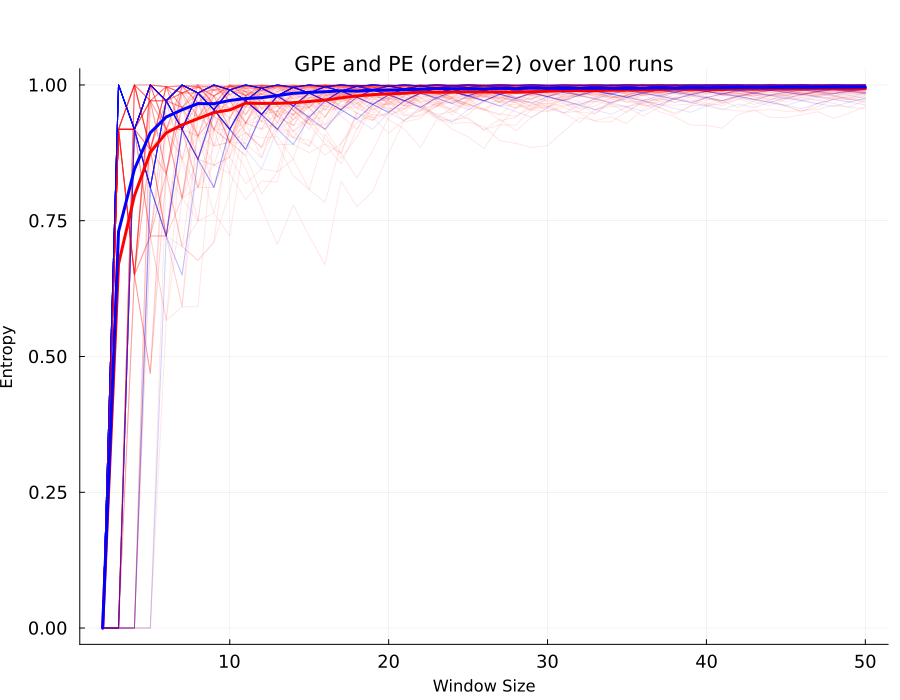}
    \includegraphics[width=0.3\linewidth]{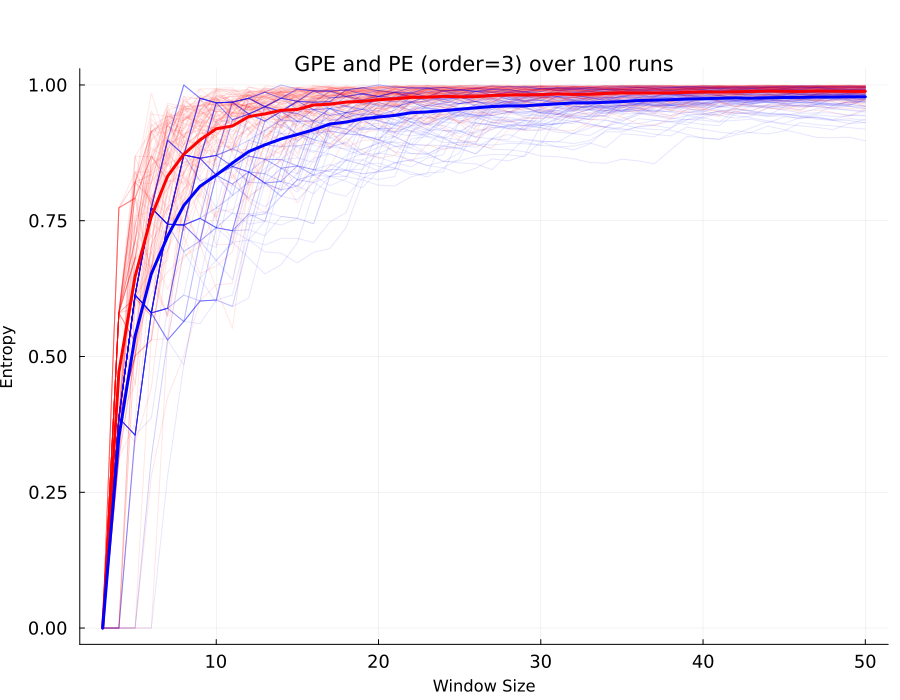}
    \includegraphics[width=0.3\linewidth]{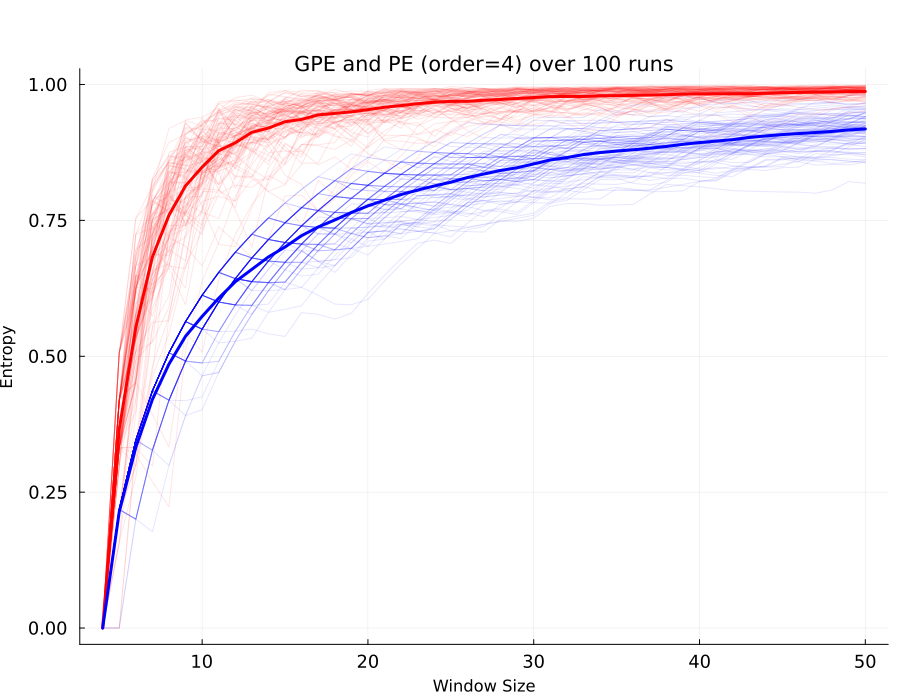}
    \includegraphics[width=0.3\linewidth]{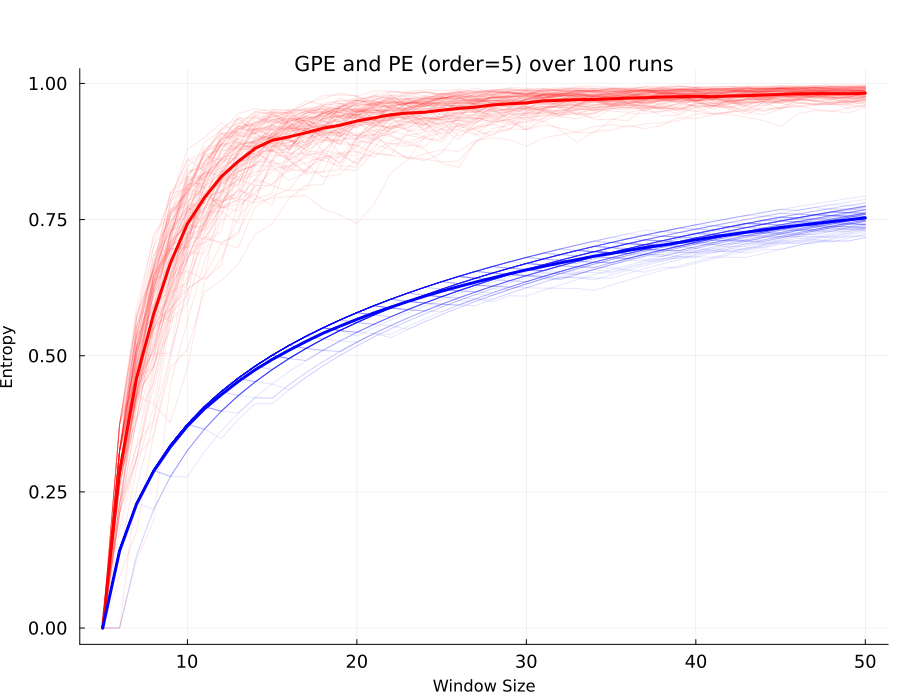}
    \hspace{0.68cm}\includegraphics[width=0.3\linewidth]{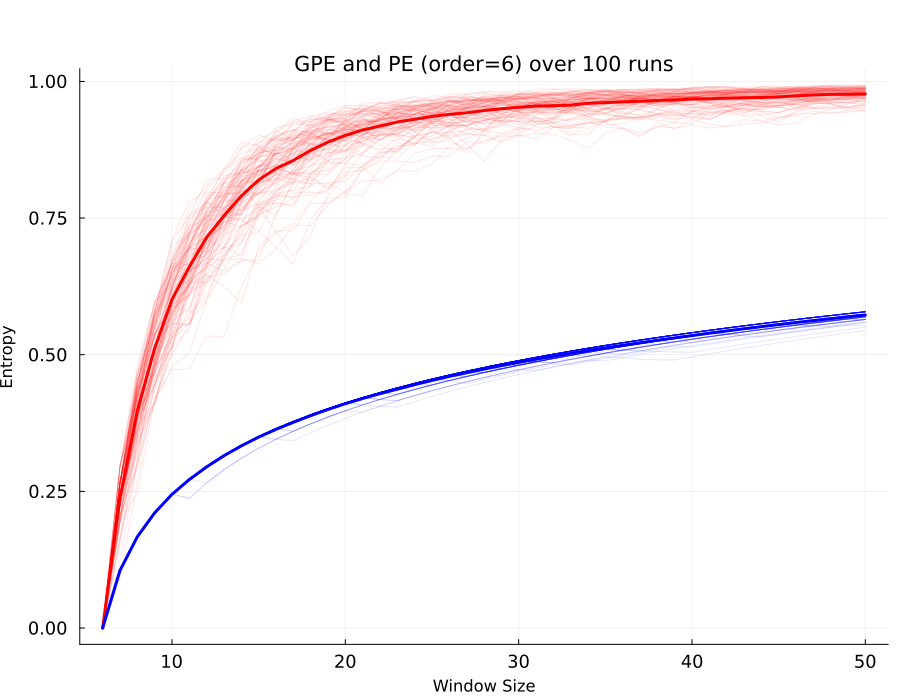}
 
    \caption{Convergence of \GPE\ (red) and \PE\ (blue) to the value $1$ in a fully random sequence.}

    \label{fig:convergence_to_1}
\end{figure}

For a given order $k$ and window length,
\GPE yields a larger sample size than \PE ,
since the former considers all possible permutation patterns within a window. We generate $100$ time series of the form \((X_t)_{t=1}^{50}\), where each $X_t$ is independently drawn from a standard normal distribution. In this fully random setting, the ``true'' (normalized) values of both \PE and \GPE are equal to 1. We compare \PE and \GPE using the same orders \(k =2,3,4,5,6\). Except for order 2, as the window size $n$ increases, \GPE converges to 1 more quickly than \PE, see Figure~\ref{fig:convergence_to_1}.


In order to show that \GPE converges faster than \PE, we also display the means of the respective $100$ realizations of the entropies.
Obviously, the convergence becomes slower for higher orders since more patterns need to be counted.

\subsection{Noise Detection}
\label{subsection:noise_detection}

We now use \(\PE\) and \(\GPE\) to detect an increase in noise within a periodic signal. Fix a period \(P \in{10,20,30}\) and a noise scale parameter \(\varepsilon \in \{0.0, 0.25, 0.5, 0.75\}\). The signal \(f(t)\) consists of two consecutive segments:

\begin{enumerate}
  \item \textbf{Less Noisy Segment} (\(1 \leq t \leq 3P\)): 
  \[
  f(t) = \sin\left(\frac{2\pi t}{P}\right) + \varepsilon \cdot \eta(t),
  \quad \eta(t) \sim \mathcal{N}(0,1),
  \]
  where noise is scaled by \(\varepsilon\), resulting in less noise.
  
  \item \textbf{More Noisy Segment} (\(3P < t \leq \frac{9}{2}P\)):
  \[
  f(t) = \sin\left(\frac{2\pi t}{P}\right) + \eta(t),
  \quad \eta(t) \sim \mathcal{N}(0,1),
  \]
  where the noise has full variance, representing an increase in noise level.
\end{enumerate}

We then generate 100 independent realizations of such signals and compute entropy values using sliding windows. We compute \(\PE\) and \(\GPE\) using orders \(k = 2, 3, 4\). We use sliding window sizes that range from $8$ to $\frac{3}{2}P+1$. For \PE, for each window size, we evaluate all feasible delay values, where the set of feasible delays is given by
\[
\mathcal{T} = \left\{1, \dots, \left\lfloor \frac{\windowsize - 1}{k - 1} \right\rfloor \right\}.
\]
Note that \(\left\lfloor \frac{\windowsize - 1}{k - 1} \right\rfloor\) represents the largest delay \(\tau\) that still allows at least one valid \(k\)-tuple within the window. For each order \(k\), we also compute the average entropy across feasible delays, \(\overline{\PE}(k)\).
In our framework, the entropy value (\PE or \GPE) serves as a classification score to distinguish less noisy observations from noisier ones. For each Monte Carlo run, we take two equal segments of length $\tfrac{3}{2}P$: a \emph{less noisy} segment (immediately before the noise onset) and a \emph{more noisy} segment (within the noisier interval). Higher entropy indicates higher noise. Varying a threshold $T$ on entropy, an observation is classified as \emph{more noisy} if $\mathrm{entropy} \ge T$ and \emph{less noisy} otherwise. True/false positives/negatives are defined as usual: 
\begin{itemize}
    \item \textbf{TP} = more noisy correctly classified,
    \item \textbf{FN} = more noisy misclassified as less noisy,
    \item \textbf{TN} = less noisy correctly classified,
    \item \textbf{FP} = less noisy misclassified as more noisy.
\end{itemize}
Sweeping $T$ over all observed entropies yields the ROC curve; the AUC is averaged over 100 Monte Carlo runs with 95\% confidence intervals.

\begin{figure}[H]
    \centering
    \begin{minipage}[t]{0.23\textwidth}
        \centering
        \includegraphics[width=\linewidth]{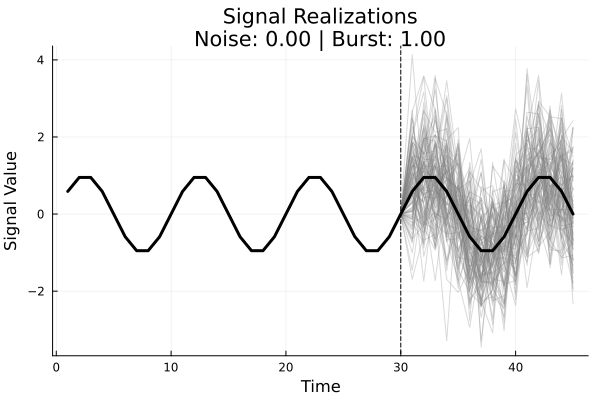}
        \label{fig:roc_15_noise0}
    \end{minipage}
    \hfill
    \begin{minipage}[t]{0.23\textwidth}
        \centering
        \includegraphics[width=\linewidth]{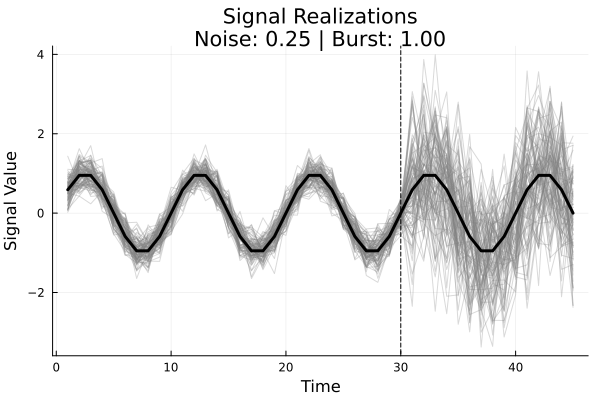}
        \label{fig:roc_15_noise025}
    \end{minipage}
    \hfill
    \begin{minipage}[t]{0.23\textwidth}
        \centering
        \includegraphics[width=\linewidth]{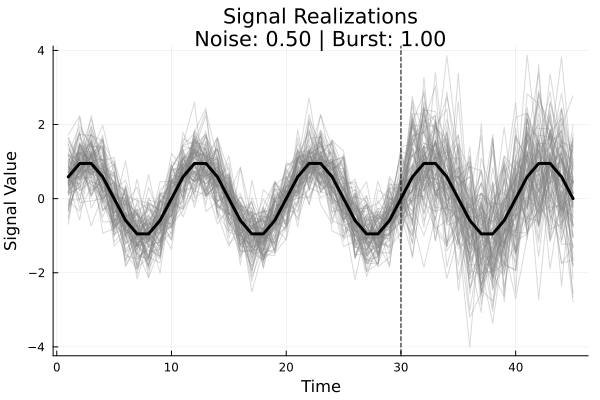}
        \label{fig:roc_15_noise05}
    \end{minipage}
    \hfill
    \begin{minipage}[t]{0.23\textwidth}
        \centering
        \includegraphics[width=\linewidth]{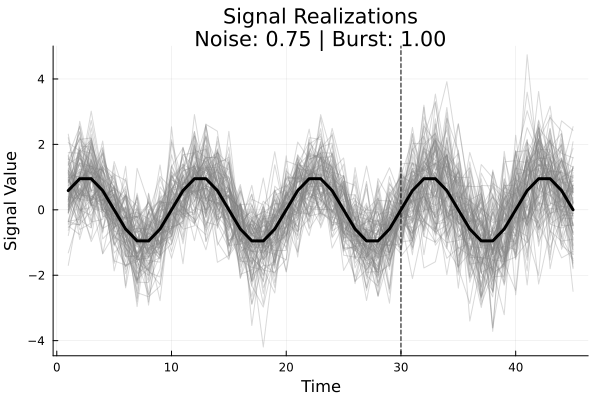}
        \label{fig:roc_15_noise075}
    \end{minipage}
    \caption{Signal realizations for varying noise levels (\(\varepsilon\)) with period 10 and length 45.}
    \label{fig:signal_realizations_15}
\end{figure}

\begin{figure}[H]
    \centering
    \begin{minipage}[t]{0.23\textwidth}
        \centering
        \includegraphics[width=\linewidth]{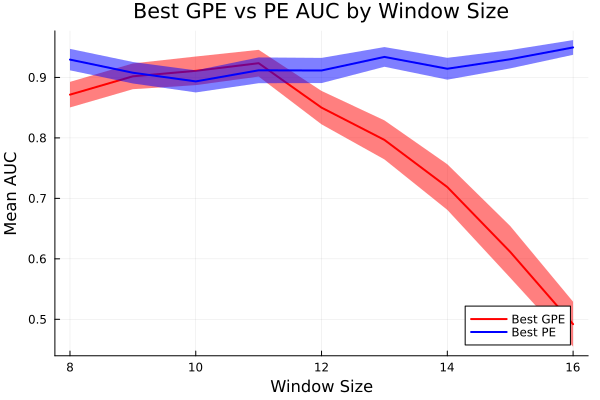}
        \caption*{(a) \(\varepsilon=0.0\)}
        \label{fig:auc_15_noise0}
    \end{minipage}
    \hfill
    \begin{minipage}[t]{0.23\textwidth}
        \centering
        \includegraphics[width=\linewidth]{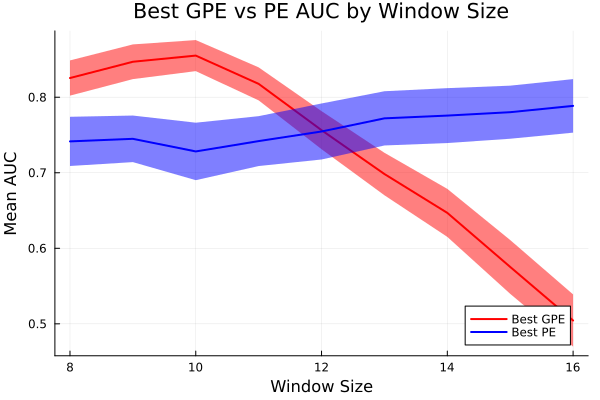}
        \caption*{(b) \(\varepsilon=0.25\)}
        \label{fig:auc_15_noise025}
    \end{minipage}
    \hfill
    \begin{minipage}[t]{0.23\textwidth}
        \centering
        \includegraphics[width=\linewidth]{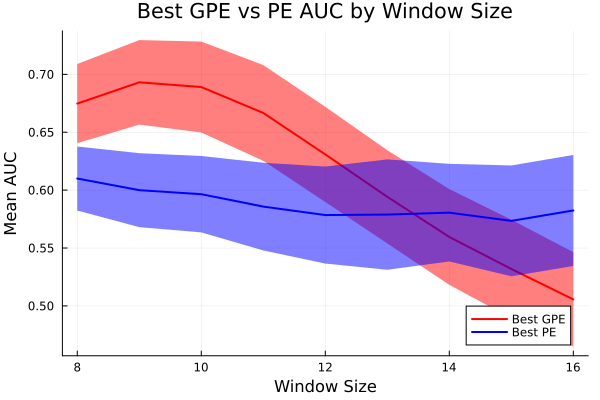}
        \caption*{(c) \(\varepsilon=0.5\)}
        \label{fig:auc_15_noise05}
    \end{minipage}
    \hfill
    \begin{minipage}[t]{0.23\textwidth}
        \centering
        \includegraphics[width=\linewidth]{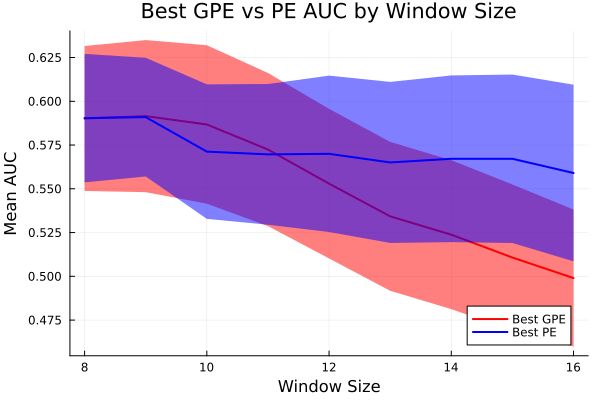}
        \caption*{(d) \(\varepsilon=0.75\)}
        \label{fig:auc_15_noise075}
    \end{minipage}
    \caption{Performance across window sizes (8 to 16) in terms of AUC for noise detection with period 10. For each window size, the highest average AUC among orders \(k = 2,3,4\) is selected for \(\GPE\). For \PE, the highest average AUC among orders \(k = 2,3,4\), feasible delays (or average across delays) is selected.}
    \label{fig:auc_performance_15}
\end{figure}

\begin{figure}[H]
    \centering
    \begin{minipage}[t]{0.23\textwidth}
        \centering
        \includegraphics[width=\linewidth]{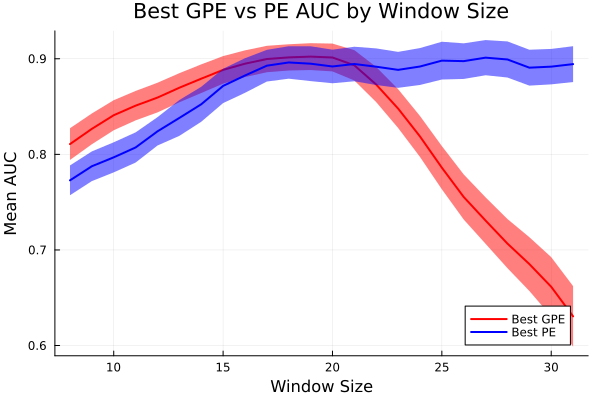}
        \caption*{(a) \(\varepsilon=0.25, P=20\)}
    \end{minipage}
    \hfill
    \begin{minipage}[t]{0.23\textwidth}
        \centering
        \includegraphics[width=\linewidth]{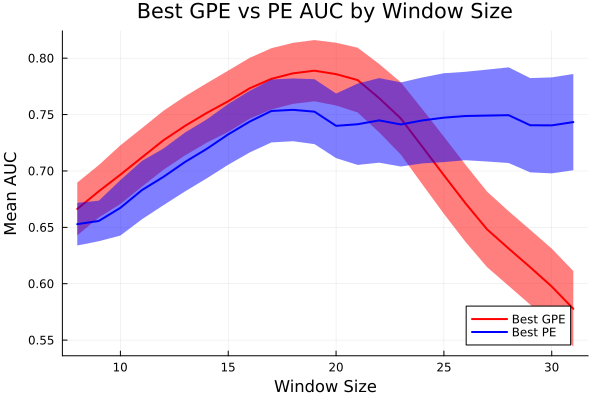}
        \caption*{(b) \(\varepsilon=0.50, P=20\)}
    \end{minipage}
    \hfill
    \begin{minipage}[t]{0.23\textwidth}
        \centering
        \includegraphics[width=\linewidth]{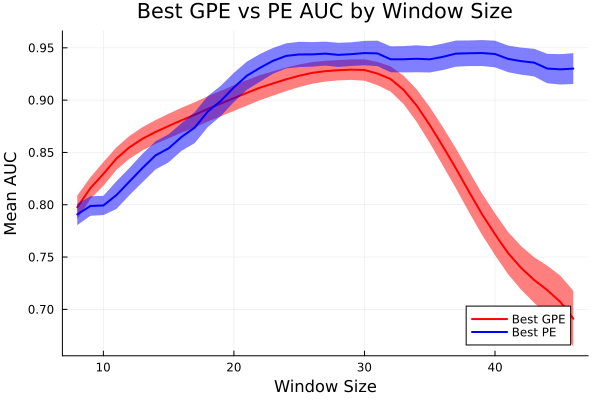}
        \caption*{(c) \(\varepsilon=0.25, P=30\)}
    \end{minipage}
    \hfill
    \begin{minipage}[t]{0.23\textwidth}
        \centering
        \includegraphics[width=\linewidth]{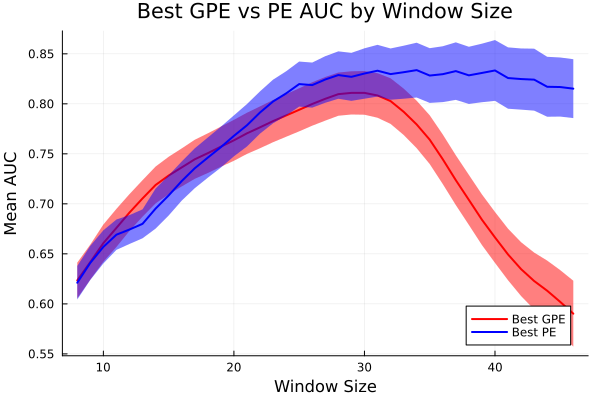}
        \caption*{(d) \(\varepsilon=0.50, P=30\)}
    \end{minipage}
    \caption{Examples with period 20 and 30}
    \label{fig:auc_performance_20_30}
\end{figure}

For $P=10$, we observe that \(\GPE\) clearly outperforms \(\PE\) for noise levels \(\varepsilon=0.25\) and \(\varepsilon=0.5\), see Figure~\ref{fig:auc_performance_15}. In the absence of noise (\(\varepsilon=0.0\)), the performances of \(\PE\) and \(\GPE\) are comparable. For the highest noise level (\(\varepsilon=0.75\)), both methods perform only slightly better than random classification. Notably, for the cases $P=20$, and $P = 30$,
 \GPE shows again optimal performance for the window size matching the signal’s period (see Figure~\ref{fig:auc_performance_20_30}). However, this time, its performance is comparable to that of \PE.

\begin{figure}[H]
    \centering
    \begin{minipage}[b]{0.35\linewidth}
        \centering
        \includegraphics[width=0.7\linewidth, trim=0 0 170 40, clip]{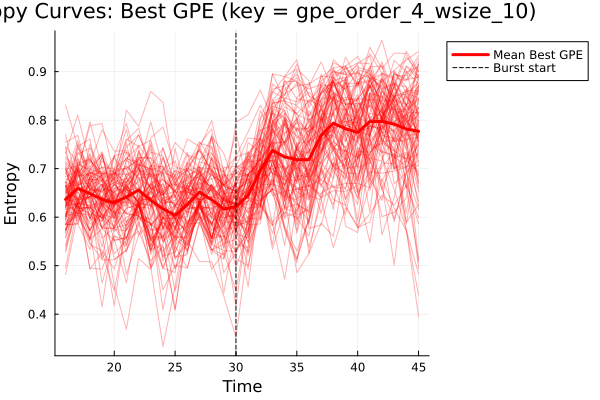} 
        \caption*{100 realizations of $\GPE(4)$, $\windowsize=10$ (average of curves in thick red)}
        \label{fig:plot_7}
    \end{minipage}\hspace{2cm}
    \begin{minipage}[b]{0.35\linewidth}
        \centering
        \includegraphics[width=0.7\linewidth, trim=0 0 170 40, clip]{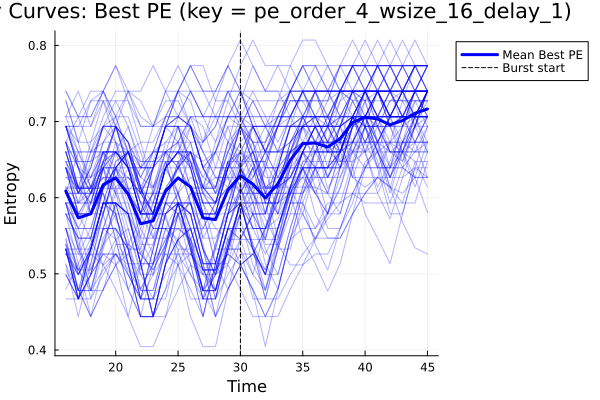} 
        \caption*{100 realizations of $\PE(4;1)$, $\windowsize=16$ (average of curves in thick blue)}
        \label{fig:plot_8}
    \end{minipage}
    \caption{Best entropies (yielding the highest average AUC) curves for $P=10$ and $\varepsilon=0.25$. Fifteen observations before the burst and fifteen within the burst.}
    \label{figure:best_entro}
\end{figure}

\subsection{Periodic signal with additive noise which increases linearly over time}
\label{subsection:periodic_additive_noise_linear_increase}

\begin{wrapfigure}{l}{0.45\textwidth}
  \includegraphics[width=.9\linewidth]{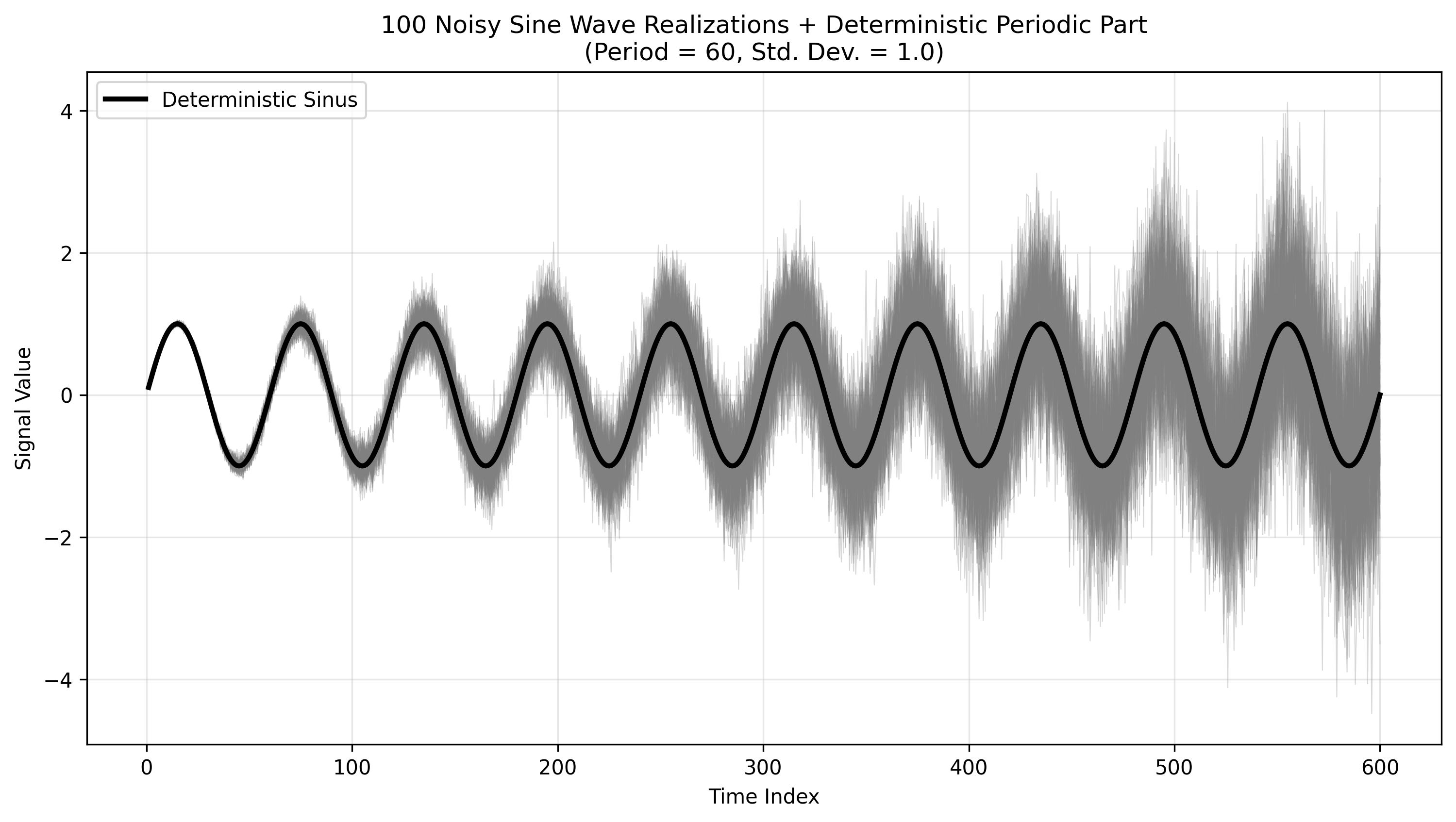}
  \captionsetup{format=plain}
  \caption{100 realizations of sinusoidal signal plus modulated noise.}

  \label{fig:sinusoidal_100}
\end{wrapfigure}
Consider the following signal
\begin{align*}
f(t) &= \sin\left( \frac{2 \pi t}{P} \right) + \frac{t}{10 P} \, \zeta(t), 
\quad \zeta(t) \sim \mathcal{N}(0, \sigma^2),
\end{align*}
where $P$ is the period of the deterministic, periodic component and $\zeta$ represents Gaussian white noise with variance $\sigma^2>0$,
with a prefactor $t/(10 P)$ that increases the standard deviation of the noise linearly over time. We consider the signal at time points $t = 1, 2, \dots, 10P$.
See \Cref{fig:sinusoidal_100} for realizations of this signal for $\sigma^2 = 1, P = 60$.
We expect any measure of entropy to reflect the increasing noise level in the signal,
and we compare how \GPE and \PE behave.

\begin{wrapfigure}{r}{0.45\textwidth}
  \includegraphics[width=.9\linewidth]{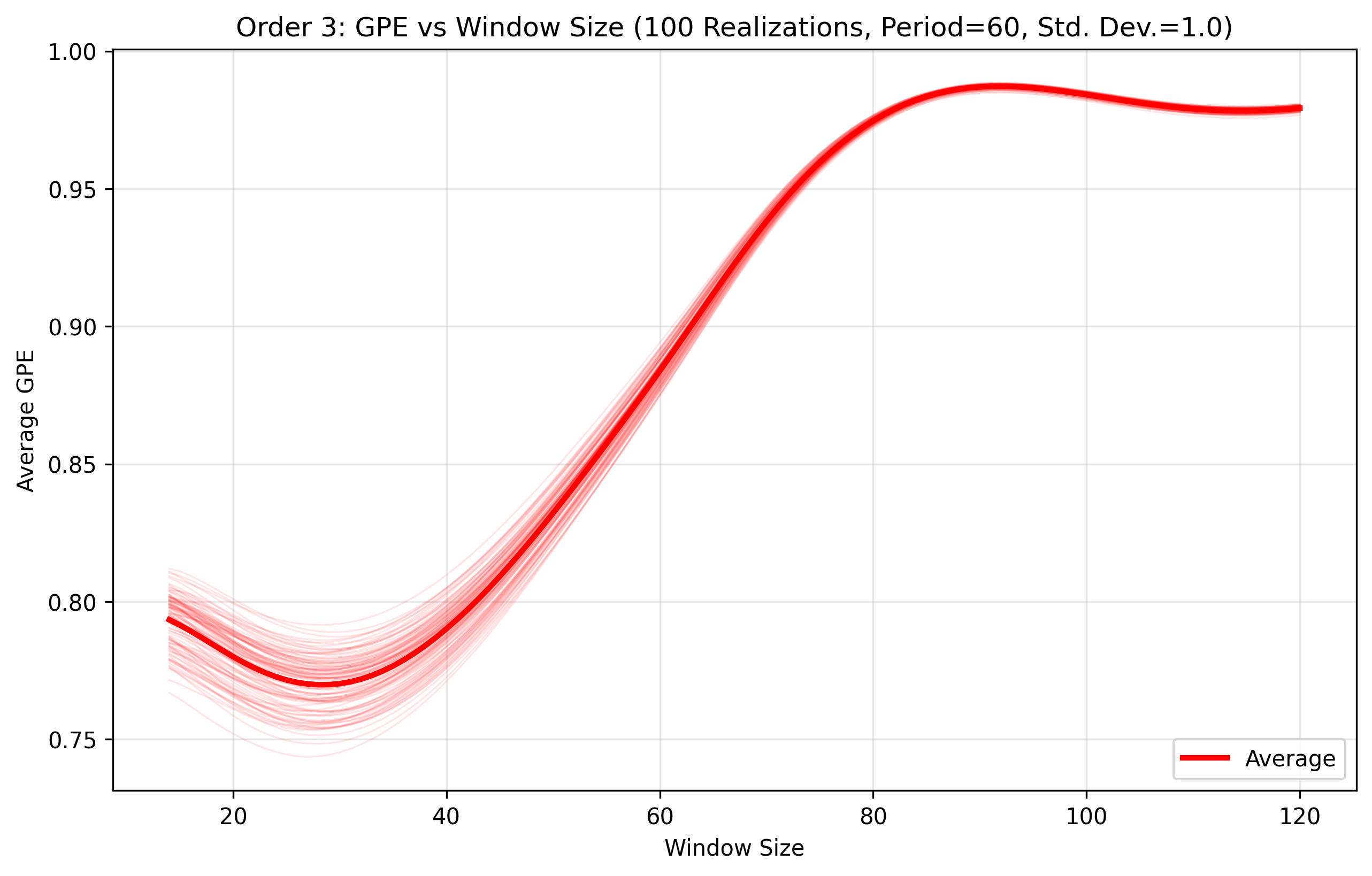}
  \captionsetup{format=plain}
  \caption{Average sliding-window entropy across different window lengths on signal shown in Figure~\ref{fig:sinusoidal_100}, computed over 100 signal realizations.}
  \label{fig:entropy_average}
\end{wrapfigure}

\textbf{Parameter selection for \(\textnormal{\GPE}\) and \(\textnormal{\PE}\)}\\
For \GPE, we use the method described at the beginning of this section
to obtain an estimate for (half) the period. As in Figure~\ref{fig:entropy_average}, it usually leads to approximately
the correct value of $30$.
A window size between $30$ and $60$ is then expected to yield
good results.

Regarding \PE, as recalled at the beginning of this section, there is no universally agreed methodology for selecting its parameters (window length, delay, etc.). We therefore evaluate \PE over multiple window sizes and delays. In addition to computing \PE with the default delay of 1, we also evaluate it using delays of 10 and 20. To further reduce sensitivity to any single delay choice, we additionally consider the average over delays in the range 1–10.

\begin{figure}[H]
    \centering
    \includegraphics[width=\textwidth]{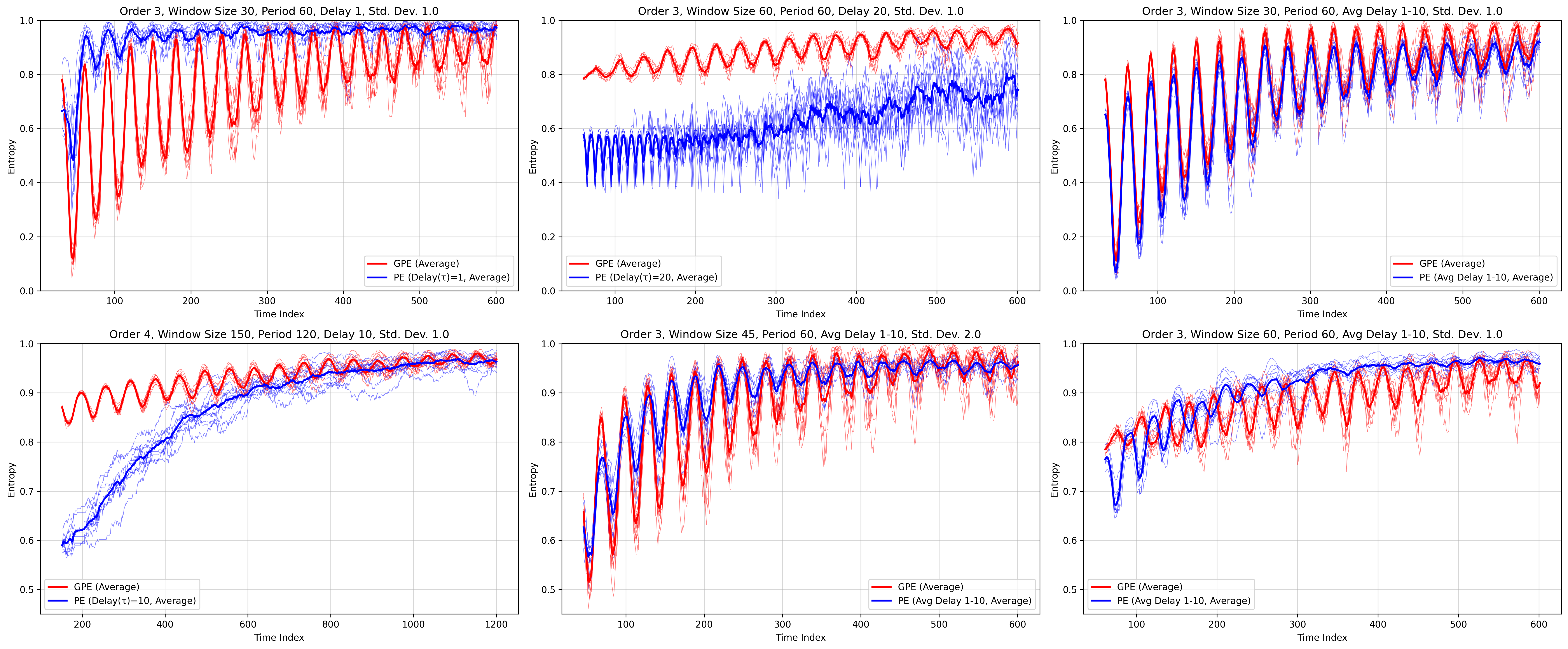}
    \caption{Signal with 100 realizations and average entropy for different window sizes, orders, period of signals, and delays.
    }
    \label{fig:Grid of plots}
\end{figure}

\textbf{Experiment}\\
In \Cref{fig:Grid of plots}, we analyze the behavior of both \PE  and \GPE under various signal conditions. The input signals consist of two distinct periods, \( P = 60 \) and \( P = 120 \), and are evaluated under two noise levels ($\sigma^{2}=1$ and $\sigma^{2}=4$).

We number the plots in \Cref{fig:Grid of plots} left to right, top to bottom.

In Plot~1,
for a window size of 30 and \( P = 60 \), \GPE exhibits a highly expressive response, capturing the underlying periodic structure effectively. In contrast, \PE with the default delay parameter \( \tau = 1 \) quickly saturates,
thereby losing sensitivity to the signal's structure.
We observe generally that \PE's performance is significantly affected by the choice of delay. For instance, with delays \( \tau = 10 \) (Plot~4) and \( \tau = 20 \) (Plot~2), \PE demonstrates erratic behavior with substantial variance across different realizations, indicating a lack of robustness.

Through extensive experimentation and fine-tuning, we found that, for this
data, \PE achieves the behavior closest to \GPE when averaging over delays in the range \( \tau = 1 \) to \( \tau = 10 \), specifically for window size $30$ and \( P = 60 \) (as shown in Plot~3).
However, this matching behavior degrades as the window size increases. In Plots~5 and 6, corresponding to window sizes $45$ and $60$ respectively (still with \( P = 60 \)), \PE again exhibits early saturation, particularly in Plot~6, whereas \GPE remains expressive and continues to reflect the underlying periodicity.

These observations highlight a key advantage of \GPE: it is more robust to changes in parameters (e.g., window size) and requires significantly less parameter tuning compared to \PE. This robustness is further confirmed in additional experiments (Plot~4 and 5) where we vary the order (e.g., to $4$), noise level (e.g., to $\sigma^{2}=4$), period (e.g., to $120$), and increase the window size to $150$ (different from the ideal window selection of $60 - 120$)---under all these conditions, \GPE consistently maintains expressive power in representing the signal structure.

\section{Conclusion and Outlook}

We have introduced Global Permutation Entropy (\GPE), a novel complexity measure for time series analysis that extends classical Permutation Entropy by considering all possible ordinal patterns of a given length within a time window, rather than only consecutive or regularly spaced patterns. This conceptually simple extension was previously computationally intractable, but recent algorithmic advances in permutation pattern counting have made its practical implementation feasible for orders up to 6.

The results presented in this paper highlight several desirable features of \(\GPE\). 

\begin{itemize}
    \item For completely random signals and fixed order \( k > 2 \), \(\GPE(k)\) converges faster than \(\PE(k)\) to the expected value of 1. This is because \(\GPE\) utilizes \emph{all possible ordinal patterns} within the window, effectively increasing the sample size for the same window length compared to \(\PE\), which only considers a tiny subset of patterns. See Section~\ref{subsection:windows_size_convergence}.
    
    \item In certain settings, \(\GPE\) also detects changes in noisy periodic signals more quickly than \(\PE\), when there is a sudden increase in noise level. We observed this for short signals, where \(\PE\) struggles due to insufficient sample size. See Section~\ref{subsection:noise_detection}.
    
    \item When gradually adding noise to a periodic signal, \(\GPE\) remains highly descriptive and robust across different orders and window sizes. Notably, \(\GPE\) performs well when the window size is chosen between half the period and one full period. This characteristic makes \(\GPE\) simpler and easier to use than \(\PE\), which requires careful tuning of delay parameters or averaging over multiple delays to achieve comparable performance. See Section~\ref{subsection:periodic_additive_noise_linear_increase}.
\end{itemize}

The larger effective sample size of \GPE compared to \PE leads to at least two consequences: the granularity of the entropy values is finer (since the ``step size'' in \eqref{eq:p_sigma} is \(\binom{\windowsize}{k}\) instead of \(\windowsize - k + 1\)), and entropy values are generally higher than for \PE (see Figure~\ref{figure:best_entro}). Indeed, in our experiments, we observe that the value of \GPE is often close to 1 but evolves on a finer scale than \PE. Furthermore, when sliding the window from time step \(t\) to \(t+1\), at most one histogram element changes for \PE, but up to \(\binom{\windowsize - 1}{k - 1}\) elements may change for \GPE. For \PE, the relative change in the histogram is either zero or exactly \(\frac{1}{\windowsize}\). In contrast, for \GPE, changes can take values from $
\left\{0, \frac{1}{\binom{\windowsize}{k}}, \ldots, \frac{k}{n}\right\}$,
since $\frac{\binom{\windowsize - 1}{k - 1}}{\binom{\windowsize}{k}} = \frac{k}{n}$. It would be interesting to explore whether the finer granularity and more dynamic behavior of \GPE could be leveraged to achieve a good performance in certain settings.

Furthermore, future work could also include:
\begin{itemize}
\item
Identifying real data sets for which \GPE performs better than \PE.

\item
Providing a fast implementation of profile 7 and therefore of $\GPE(7)$. This is already feasible, relying on results from \cite{beniamini2024counting}.

\item
Permutation entropy considers only consecutive or regularly spaced patterns.
Global permutation entropy considers all patterns (also known as the classical permutation patterns).
What other types of patterns could be considered to define a permutation entropy?

\item Consider a measure of entropy based on corner trees as follows.  
For a fixed order $k$, consider the maximal set of corner trees with $k$ vertices that span linearly independent directions of patterns.  
Let $\{T_1,\dots,T_m\}$ denote these trees, and let $c_i$ be the number of occurrences of $T_i$ in the ranked time series, with total 
$C = \sum_{i=1}^m c_i$. Define the empirical distribution $p_i = c_i/C$.  
The raw \emph{corner-tree entropy} is given by
\[
\rawCTPE(k) \;:=\; -\sum_{i=1}^m p_i \log p_i.
\]
Note that, although from level~4 onward the corner trees no longer span the entire profile, this entropy can be evaluated in just $\mathcal{O}(n \log n)$ time.

\item
Applying known extensions of \PE to \GPE, such as conditional entropy \cite{unakafov2014conditional}.

\end{itemize}

\printbibliography
\end{document}